\documentclass[runningheads]{llncs}
\usepackage[T1]{fontenc}
\usepackage{amsmath}
\usepackage[normalem]{ulem}

\usepackage{hyperref,cleveref}
\usepackage{booktabs}

\usepackage{algorithm2e}
\usepackage[inline]{enumitem}
\usepackage{subfiles}
\usepackage{commands}

\usepackage{multirow}
\usepackage{subcaption}
\usepackage{graphicx}

\title{Complex System Exploration\\ with Interactive Human Guidance}

\author{
    Bastien Morel\inst{1} \and
    Clément Moulin-Frier\inst{2} \and
    Pascal Barla\inst{2} 
}

\authorrunning{B. Morel et al.}

\institute{
    Univ. Bordeaux, CNRS, Bordeaux INP, LaBRI,\\ UMR 5800, F-33400 Talence, France\and INRIA, France\\
    \email{firstname.lastname@inria.fr}
}

\begin{document}

\maketitle
\begin{abstract}
The diversity of patterns that emerge from complex systems motivates their use for scientific or artistic purposes.
When exploring these systems, the challenges faced are the size of the parameter space and the strongly non-linear mapping between parameters and emerging patterns.
In addition, artists and scientists who explore complex systems do so with an expectation of particular patterns.
Taking these expectations into account adds a new set of challenges, which the exploration process must address.
We provide design choices and their implementation to address these challenges; enabling the maximization of the diversity of patterns discovered in the user's region of interest -- which we call the constrained diversity -- in a sample-efficient manner.
The region of interest is expressed in the form of explicit constraints. These constraints are formulated by the user in a system-agnostic way, and their addition enables interactive system exploration leading to constrained diversity, while maintaining global diversity.
\end{abstract}
\vspace{-1cm}
\begin{figure}[h!]
    \centering
    \includegraphics[height=4cm]{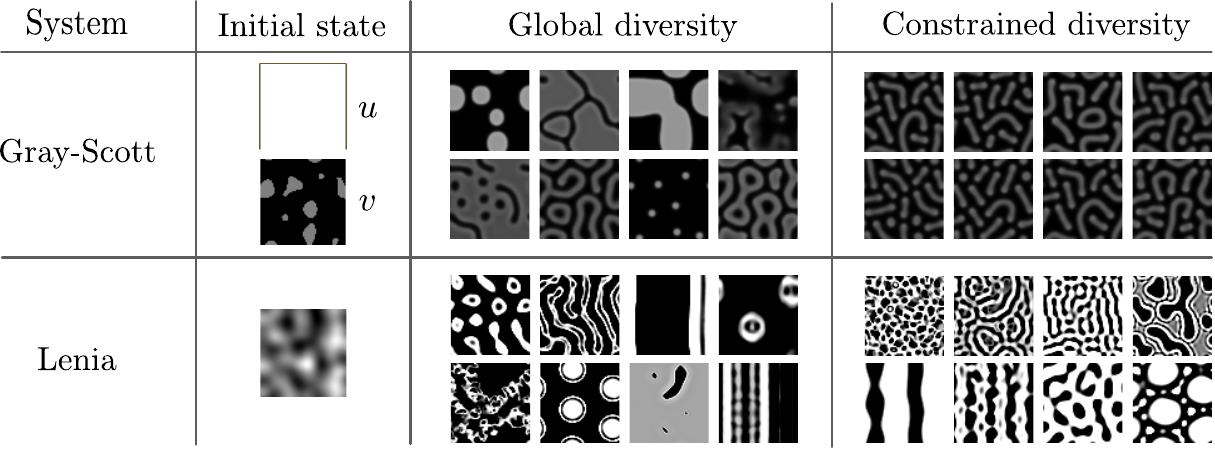}
    \caption{\textbf{Our method explores the parameter space of complex systems to discover patterns that belong to the user's region of interest.} Compared to global diversity search, our constrained diversity search approach allows the exploration process to be guided toward user-specific patterns. In this example, the user expects to obtain images with a dark area proportion belonging to $[0.3, 0.4]$.}
    \label{fig:patterns_preview}
\end{figure}

\section{Introduction}
\label{sec:intro}
Complex systems are prone to the self-organization of spatio-temporal patterns from the local interaction of their constituents.
In nature, many patterns result from the dynamics of such complex systems \cite{ball_self-made_1998}, which has motivated the artificial life community to rely on computer simulations in order to understand and explore such phenomena \cite{dorin_what_2024}, as models of natural phenomena, or for generative arts.
Taking advantage of complex system, requires to efficiently explore the space of patterns they can generate.
Exploring an artificial complex system involves sampling the parameter space that defines it, with the aim of finding a specific pattern or discovering a set of diverse new patterns.
Searching for patterns produced by a complex system requires dealing with excessively large, potentially open-ended search spaces \cite{etcheverry_curiosity-driven_2023}, combined with a strongly non-linear mapping between system's parameters and system's behaviors.
Random sampling in parameter space is inefficient compared to machine learning methods based on optimization or diversity. 
In particular, intrinsically motivated goal exploration process (IMGEP) \cite{reinke_intrinsically_2020} tends to maximize the diversity of discovered patterns in behavior space, i.e., a representation space characterizing the set of possible patterns in a given system (more details are given in Section \ref{sec:background}).

In addition to a large/open-ended search space, exploration can take place with particular expectations in the user's mind, and feedback is needed to provide them.
Adding a human to the exploration loop forces the exploration process to take user constraints into account.
The goal is then to achieve what we call a \emph{constrained diversity} of patterns (see Figure~\ref{fig:patterns_preview}).
Indeed, of all the patterns that can be generated by a complex system, the user may only be interested in a subset, which we term a Region Of Interest (ROI).
But as the system is explored, this ROI may vary, as the user takes advantage of the serendipity offered by exploration.
In addition, not all points in the ROI may be accessible to the complex system.

\begin{figure*}[!h]
    \centering
    \includegraphics[width=0.9\textwidth]{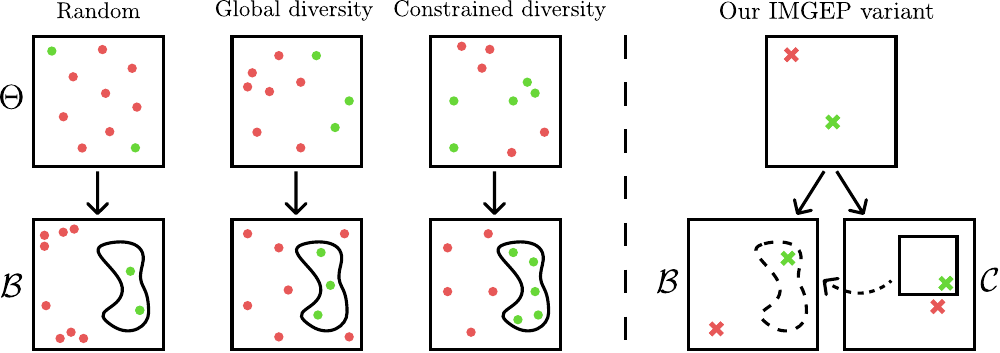}
    \caption{\textbf{Overview of our approach.}
    Our goal is to define a procedure for exploring the parameter space $\paramspace$ of a complex system, taking into account the target user's ROI (closed region) in a behavior space $\repspace$.
    A random sampling in $\paramspace$ leads to a poor diversity in $\repspace$.
    Diversity-based methods usually achieve a high global diversity.
    However, most of the samples end up outside the ROI.
    Our goal is to find a sampling method that maximizes constrained diversity (i.e., resulting in more samples inside the user's ROI and whose diversity is maximized).
    To this end, we introduce an IMGEP variant that relies on an additional  constraint space $\constraintspace$ in which the ROI is more easily expressible than in $\repspace$.}
    \label{fig:exposition}
\end{figure*}

In this work, our goal is to put the human right into the loop of complex system exploration. 
An overview of our approach is given in Figure~\ref{fig:exposition}.
In particular, we seek \emph{interactive} feedback, meaning the exploration loop should be fast enough for the user to adapt their ROI at each iteration.
In addition, we want our method to be readily usable without any prior knowledge on the system behavior.
We expose design choices that allow for such an interactive exploration process of complex systems, taking into account potentially varying ROI.
We implement those design choices through a modified IMGEP algorithm, and we show the ability of our implementation to maximize constrained diversity through quantitative and qualitative results.

Lastly, we discuss merits and flaws of those design choices and we identify avenues of future work.

\section{Related Works}

    \label{sec:relatetedworks}
    Optimization-based methods, such as evolutionary algorithms \cite{mitchell_evolving_1996} or gradient descent \cite{mordvintsev_growing_2020}, can be used to find parameters that lead to a pattern with \emph{a priori} optimal characteristics.
    What characteristics are considered as optimal has to be defined by the user in the form of a scalar objective function of the parameter space.
    Optimization-based methods are subject to standard issues in optimization, in particular deception in the presence of strong local optima.
    
    Diversity-based methods aim to find sets of parameters that lead to a diversity of patterns, in a so-called behavior space.
    Intrinsically motivated goal exploration process (IMGEP) \cite{forestier_intrinsically_2017,reinke_intrinsically_2020} belongs to this class.
    However, defining a behavior space a priori can potentially limit the global diversity of the discovered patterns.
    To circumvent this problem Etcheverry et al. \cite{etcheverry_hierarchically_2020} introduce the use of a hierarchy of behavioral spaces, a hierarchy built using a \emph{variational autoencoder} (VAE) \cite{kingma_auto-encoding_2013}, which implies a high calculation cost.
    
    In order to benefit from the advantages of diversity-based and optimization-base methods, hybrid methods have focused on finding parameters that lead to a diversity of patterns with a priori optimal characteristics.
    MAP-Elites \cite{mouret_illuminating_2015} is a method that tends to maintaining a population (set of parameters) with the highest fitness scores. The behavioral space is divided into sub-sections, and the algorithm seeks to find for each sub-section the parameters with the highest fitness scores.
    Like IMGEP, the result of MAP-Elite depends on the a priori formulation of the behavior space.
    AURORA \cite{grillotti_unsupervised_2022} relaxes this constraint and allows the use of a behavior space that is not hand-coded.
    IMGEP instances can also be defined to benefit from gradient descent. Hamon et al. \cite{hamon_discovering_2024} take advantages of curriculum learning capabilities and enable the discovery of increasingly complex behaviors through the use of an optimization process inside the parameter sampling strategy. 
    
    In this work, we aim to place the user in the exploration loop as a guide, without any a priori knowledge of the system or the mapping of parameters to behavior.
    IMGEP-HOLMES \cite{etcheverry_hierarchically_2020} enables such user guidance, by considering a hierarchy of behavior spaces and allowing the user to regularly rate each of them, thereby biasing future exploration with user feedback.
    However, the implicit definition of the ROI (through user rating) and the computational cost of updating the hierarchy implies a tradeoff between sample efficiency and ROI exploration speed.
    OMNI \cite{zhang_omni_2024} enables open-ended learning in a vast research space,  with the “interesting” tasks to be learned determined by a foundation model.
    The ROI is therefore modeled by a foundation model, but is difficult to customize and is built from existing data.
    ASAL \cite{kumar_automating_2024} also proposes the use of foundation models for automated discovery, notably to maximize global diversity (illuminate the behavior space) or find parameters corresponding to a target (image or prompt).
    The core of these methods relies on having a human agent give instructions on the direction the discovery should take.
    However, these methods are positioned in a non-interactive context, not allowing rapid redirection of the search when the user changes his ROI.

\section{Design choices}
\label{sec:design}
In this section, we give the desiderata induced by the interactive addition of the human in the loop, followed by design choices made to satisfy these desiderata.
\subsection{Desiderata}

\begin{enumerate}[label=\textbf{D\arabic*}, ref=D\arabic*]
    \item \label{desideratum:diversity} \textbf{Constrained diversity}
    Matching the user's ROI to the system's behavior space is not always feasible.
    Hence, to ensure that the ROI has been effectively explored thoroughly, it is necessary that the diversity found within the ROI at the end of the exploration be maximal; we call this diversity, the constrained diversity.
    \item \label{desideratum:agnostic} \textbf{System-agnostic}
    We place exploration in a context in which the user does not necessarily have any prior knowledge of the parameter-behavior mapping of the system being explored.
    The exploration process must therefore be agnostic to the complex system being explored, and must not be based on data from the system prior to exploration, aside from the type of generated pattern.
    \item \label{desideratum:feedback} \textbf{User feedback}
    As the user can change his mind and adjust his ROI during exploration,
    at any point during the exploration, the user must be able to steer the exploration in a direction that suits him best.
    \item \label{desideratum:interactivity} \textbf{Interactive exploration}
    The fact that the user can give feedback to adapt the algorithm's behavior implies a fast response time for the algorithm, so that the user can determine whether his guidance is relevant or not.
    This means that the exploration loop must be executed in interactive time.

\end{enumerate}

\subsection{Choices}

\begin{enumerate}[label=\textbf{C\arabic*}, ref=C\arabic*]

\item \label{choice:constraints} \textbf{Explicit constraints}
    We let users provide feedback on exploration in the form of explicit constraints on pattern properties (\ref{desideratum:diversity}). 
    Sampled patterns are thus labelled as inliers if they lie within the ROI, and outliers otherwise, independently of the choice of system (\ref{desideratum:agnostic}).
    Users may also change their mind through modifications of the ROI during exploration (\ref{desideratum:feedback}).
    
\item \label{choice:diversity} \textbf{Maintaining global diversity}
    We propose to not only maximize constrained diversity, but also global diversity, with two objectives in mind.
    First, it should give the user a better overview of the patterns attainable by the system, as this is not known in advance (\ref{desideratum:agnostic}).
    Second, having a diverse set of samples (inliers or outliers) may also turn out to be useful whenever the user changes their ROI during the exploration process (\ref{desideratum:interactivity}).

\item \label{choice:behavior-space} \textbf{Non-specific behavior space}
    We construct the behavior space from a set of non-system-dependent features, which has two benefits :
    it does not require a priori knowledge on the system (\ref{desideratum:agnostic}); neither does it require any costly a posteriori analysis of system outputs, such as VAE training (\ref{desideratum:interactivity}).
    This often results in a high-dimensional behavior space, which requires special care.
\end{enumerate}

\section{Background}
\label{sec:background}
\subsection{Complex System}

Formally, in this paper, a system is a tuple $\mathcal{S} = (\statespace, U, \paramspace)$, where $A$ is the state space, $U$ the update rule and $\paramspace$ the parameter space.
Following Ladyman et al.~\cite{Ladyman2013}, a \emph{complex} system is a system composed of many entities that interact with one another.

We consider $n_s$ entities each having a set of states of dimension $d_s$.
The $n_s \times d_s$ dimensional space $\statespace$ corresponds to the state space of the system.
At each time step, the state of the system is updated following the update rule $U : \statespace \times \paramspace \rightarrow \statespace$.
The update rule is parameterised by the current state of the system $A_t$ and a parameters $\param \in \paramspace$.

A rollout $R_{0\rightarrow T}$ of a system $\mathcal{S}$, with the initial state $\state_0 \in \statespace$ and parameterised by $\param \in \paramspace$, is the sequence of states $R_{0\rightarrow T}(\mathcal{S}, \state_0, \param) = \state_0 \rightarrow \state_1 \rightarrow ... \rightarrow \state_T$ with each obtained by the iterative application of the update rule $\state_t = U(\state_{t-1}, \param)$.

An observation of a rollout is obtained by applying a function $O : \statespace^T \rightarrow \obsspace$, where $\obsspace$ is the observation space.
In the general case, observation is defined as the sequence of states obtained by the rollout (i.e., $\obsspace := \statespace^T$).
In our work, we use the word \emph{observation} and \emph{result} interchangeably, since observation is what is shown to the user as the result of a rollout.

\subsection{Intrinsically Motivated Goal Exploration Process (IMGEP)}

In this article, the initial state is not varying from one rollout to another of the same system and the result proposed to the user corresponds only to the last state of the rollout, therefore for sake of clarity $R_{0\rightarrow T}(\mathcal{S}, \state_0, \param) := R_{0\rightarrow T}(\mathcal{S}, \param)$ and $O(R_{0\rightarrow T}(\mathcal{S}, \param)) := \state_T$.

We implement our design choices through an Intrinsically Motivated Goal Exploration Process (IMGEP).
Originally  developed  to  model curiosity-driven learning in robotics \cite{forestier_intrinsically_2017}, IMGEP has been transposed to complex system’s diversity search \cite{etcheverry_curiosity-driven_2023}.

In this formulation, the aim of an IMGEP is to maximize diversity found in the behavior space $\repspace$ (a dimensionaly-reduced space that summarizes observations and emphasizes diversity), in a sample-efficient manner.
To do this, an IMGEP carries out a sequence of experiments, each corresponding to a rollout of the system $R_{0\rightarrow T}(\mathcal{S}, \param_i)$, with $i$ the $i$-th experiment.
The parameter $\param_i$ of all experiments, their corresponding results $\observation_i$ and behavior $\representation_i$ are stored in a history $\history = \{(\param_i, \observation_i, \representation_i), \forall i \in [1, N]\}$ where $N$ is the number of experiments.
The diversity produced by an exploration is then measured from the set of behavior space points contained in the history.

One of the key principles of an IMGEP is the reuse of past experiments.
In the case of diversity search, the value of reusing experiments is illustrated by the fact that parameters leading to a specific point in behavior space can be identified more quickly when points close to it have already been found.

To integrate this principle, the process is divided in 4 steps:
1) Sampling a goal, according to a distribution parameterized by the history and the behavior space, $\goal_k \sim \goalsampling(\history, \repspace)$;
2) Inferring a parameter to try to reach this goal, inference is performed on the basis of a parameter sampling policy $\param_k \sim \policy(\goal_k, \history, \repspace)$;
3) Executing a rollout of the system $R_{0\rightarrow T}(S, \param_k)$ and getting the observation $\observation_k$;
4) Getting the behavior $\representation_k$ of the observation $\observation_k$, and saving the experiment outcomes in the history: $\history \leftarrow \history \cup \{(\param_k, \observation_k, \representation_k)\}$.
We refer the interested reader to \cite{reinke_intrinsically_2020,etcheverry_hierarchically_2020} for specific examples of $\goalsampling$ and $\policy$.

\section{Our method}
\label{sec:method}
Recall that our aim is to achieve interactive exploration of a complex system guided by user's feedback without any prior knowledge, while achieving maximal constrained diversity (see Figure~\ref{fig:exposition}).
We implement our design choices through a variation of an IMGEP.
In accordance with these choices, we constitute the behavior space $\repspace$ with a set of non-system specific features (\ref{choice:behavior-space}).
Such representation space irremediably biases the diversity produced by the process, as shown in \cite{reinke_intrinsically_2020}.
In the following, we describe how we countermeasure this problem.

\subsection{Augmented history}

To capture the user's ROI and guide the exploration towards a specific subspace, we augment the history with classification values that emphasize the validity of the experiment in relation to the user's ROI.
The history in our IMGEP variant is thus defined as $\history := \{(\param_i, \observation_i, \representation_i, \validity_i), \forall i \in [0, N]\}$, where $c_i \in \{-1, 1\}$ corresponds to an outlier or inlier classification.
Those classifications are made based on constraints formulated explicitly by the user (\ref{choice:constraints}), which are constructed from features extracted from the observation space.
The selected features are chosen to be intuitive and easily understandable by the user.
For instance, one may use coarseness, directionality or contrast of Tamura's image features \cite{tamura_textural_1978}.

The user's ROI may thus now be seen as a simple region in a constraint space $\constraintspace$.
As illustrated in Figure~\ref{fig:exposition}-right, $\constraintspace$ is usually different from the behavior space $\repspace$.
Indeed, constraints are only used to classify samples as inliers or outliers, they are not involved in any other calculation during the exploration process.
As a result the boundaries of the ROI in behavior space $\repspace$ are not explicitly available, but this is not an issue for the IMGEP variant presented in the next section.
In addition, whenever the user makes changes to the ROI during the exploration process, the exploration process does not need to be restarted since we only need to modify the classification values $c_i$ of all entries in the history. 

\subsection{Balanced IMGEP policy}
\label{sec:balanced-policy}

Now that we have incorporated classification values to the history, it is possible to balance the parameter selection policy $\policy$ towards the ROI.
To do so, we first sample a goal $\goal \sim \goalsampling(\history, \repspace)$, where $\goalsampling$ is a uniform distribution of the hyper-rectangle including all behavior points found during experiments, i.e., all $\representation_i$ in $\history$ regardless of their classification.
The basic nearest neighbor policy $\policy$ works in two steps.
It first identifies the parameter $\param_k$ that leads to the behavior $\representation_k$ closest to the goal $\goal$ in the history $\history$, with  $\representation_k = \arg\min_{\representation_i} \Vert g - \representation_i \Vert$.
It then applies a mutation to it (for more details, see~\cite{reinke_intrinsically_2020,etcheverry_hierarchically_2020}). 
This policy needs to be adjusted in our case in order to maximize constrained diversity.
We first want to take explicit constraints into account (\ref{choice:constraints}) while maintaining a global diversity (\ref{choice:diversity}), as the ROI might be subject to change during exploration.
Our solution consists of introducing a balance between global diversity samples and constrained diversity samples using a meta-parameter, which is used as the probability of a Bernouilli dristibution (we use a probability of $0.5$ in practice).
This distribution is sampled at each experiment to determine the subset of behaviors used to determine the nearest point: in case of a global diversity sample, all entries in the history are used regardless of their classification value; in case of a constrained diversity sample, only entries corresponding to inliers (i.e., with $c_i = 1$) are taken into account.

Another difference in our case is that the behavior space $\repspace$ is not specific to the explored system (C3), which raises two issues.
First, the dimensions of $\repspace$ might not be commensurate. 
We thus standardize both the history $\history$ and goal $\goal$ for each dimension independently, at each exploration step.
Second, $\repspace$ might be high-dimensional, and our approach be subject to the curse of dimensionality: the nearest neighbor search might then 
pick neighbors mostly in a small subset of dimensions, hindering diversity.
We mitigate that issue by using a subset of dimensions in $\repspace$, which are sampled uniformly.
The number of sub-dimensions is a meta-parameter of our process (we use $3$ sub-dimensions in practice).

\section{Results}
\label{sec:results}
In this section, we present results obtained with the implementation of our IMGEP variant.
For this purpose, two complex systems are studied.
First, Gray-Scott, a reaction-diffusion system that comprises a search space with a small sub-region producing patterns; in such search spaces, random sampling is largely inefficient.
Second, Lenia, a class of continuous cellular automata, in which the parameter space is high-dimensional.

\subsection{Methodology}

For each studied system, 4 exploration methods are tested :
\begin{itemize}
    \item \textbf{Random (R)} : a random sampling of the parameter space.
    \item \textbf{Nearest (N)} : IMGEP exploration method, with nearest neighbors computed on the whole behavior space.
    \item \textbf{NearestRandomAxes (NRA)} : IMGEP exploration same as N, except that distances are computed in a 3-dimensional subspace of $\repspace$.
    \item \textbf{NearestRandomAxesBalanced (NRAB)} : our IMGEP variant with the full balanced policy of Section~\ref{sec:balanced-policy}.
\end{itemize}
All three IMGEP methods have their $N_{init}$ first samples chosen with the Random method.
The mutation applied to the candidate parameters follows a normal distribution centered at $0$, with a standard deviation $\sigma_m$ given in the caption of Figure~\ref{fig:plots}.
The 10-dimensional behavior space $\repspace$ is constructed from Hu moment invariants \cite{hu_visual_1962}, the mean value of pixels, and the volume (i.e the proportion of pixels with a value greater than a given threshold $\epsilon = 1e-5$).
The ROI of the user is explicitly given by a constraint on the volume of the observation. To satisfy this constraint, the volume must be in the range [0.6, 0.7].

\subsection{Gray-Scott Reaction-Diffusion}
\label{subsec:gray_scott}
Reaction-Diffusion are complex systems involving different chemical concentrations.
Each concentration has the ability to diffuse into the medium and react with other concentrations or itself.
Such systems were initially introduced by Turing \cite{turing_chemical_1952}, and are prone to the emergence of so-called Turing patterns (see Figure~\ref{fig:patterns_preview}.a).
The Gray-Scott reaction-diffusion \cite{gray_autocatalytic_1984} is composed of two concentrations $u$ and $v$.
The kinematic of the reaction is given by the following equations :
\begin{equation}
    \begin{aligned}
        \frac{\partial u}{\partial t} & = D_u \nabla^2u - uv^2 + f(1 - u) \\
        \frac{\partial v}{\partial t} & = D_v \nabla^2v + uv^2 - (f - k)v \\
    \end{aligned}
    \label{eq:gray_scott}
\end{equation}

In the experiment, the update rule $U$ is the forward Euler integration of the equations \ref{eq:gray_scott} on a discrete toroidal grid of resolution $32 \times 32$. 
Diffusion terms are fixed to $D_u = 0.5$ and $D_v = 0.25$.
The parameter space is then composed of the feeding-rate $f$ (the quantity of $u$ constantly added) and the killing-rate $k$ (the quantity of $v$ constantly removed).
The initial concentration of $u$ is set to 1 and a Perlin noise \cite{perlin_image_1985} thresholded at $0.7$ gives the concentration of $v$.

Due to the nature of the equations, the pattern emerging in the $u$ channel is almost similar to that emerging in the $v$ channel.
Therefore, the observation is constructed solely from the values of the $u$ channel.

\subsection{Lenia}
\label{subsec:lenia}

Lenia is a class of continuous cellular automaton \cite{chan_lenia_2019,chan_lenia_2020} mainly studied for its ability to produce spatially-localized patterns.
With non-locally distributed initial conditions, the system tends to produce Turing like patterns, as illustrated in the Figure~\ref{fig:patterns_preview}.b.
The update rule of the automata is the following:
\begin{equation}
    A_{t + 1} = A_t + dt\sum_{k} h_{k} \cdot G_k(K_k * A_t)
\end{equation}
The growth function $G_k$ is a gaussian parametrised by $\mu_k$ and $\sigma_k$, $G_k(x) = 2\ \mathrm{exp}\left(-\frac{(x - \mu_k)^2}{2\sigma_k^2}\right) - 1$.
Each kernel is made of $n$ overlapping gaussian bumps.
Our implementation is based on a differentiable version of Lenia \cite{hamon_discovering_2024} which defines kernels as $K_k(x) = \sum_{i}^{n}b_i\ \mathrm{exp}\left(-\frac{\left(\frac{x}{r_kR} - a_i\right)^2}{2w_i^2}\right)$.

We simulate a Lenia rollout on a discrete toroidal grid of resolution $64 \times 64$.
The system can be set up with several channels, but in our case we use only one.
The number of kernels is fixed to $k = 3$, each of them having $n = 3$ bumps.
Therefore the number of parameters is : $3(3 \times 3 + 4) + 2 = 41$.
The initial state is set by a Perlin noise \cite{perlin_image_1985} (not thresholded in this case).

\begin{figure*}[h!]
    \centering
    \vspace{-4mm}
    \begin{subfigure}[h]{0.48\textwidth}
        \centering
        \subcaption{Global diversity Gray-Scott}
        \includegraphics[width=\linewidth]{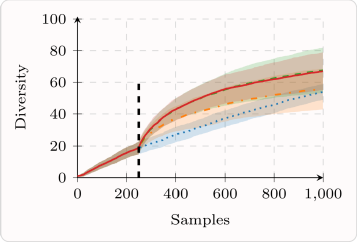}
    \end{subfigure}
    \hfill
    \begin{subfigure}[h]{0.48\textwidth}
        \centering
        \subcaption{Constrained diversity Gray-Scott}
        \includegraphics[width=\linewidth]{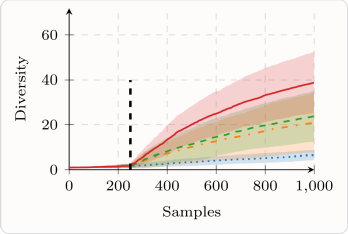}
    \end{subfigure}
    \vfill
    \begin{subfigure}[h]{0.48\textwidth}
        \centering
        \subcaption{Global diversity Lenia}
        \includegraphics[width=\linewidth]{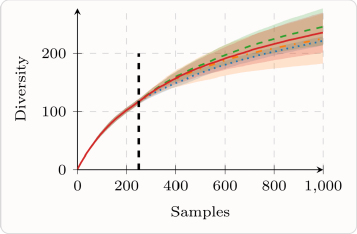}
    \end{subfigure}
    \hfill
    \begin{subfigure}[h]{0.48\textwidth}
        \centering
        \subcaption{Constrained diversity Lenia}
        \includegraphics[width=\linewidth]{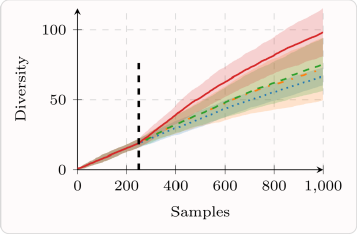}
    \end{subfigure}
    \vfill
    \begin{subfigure}[h]{\textwidth}
        \centering
        \includegraphics[width=0.5\linewidth]{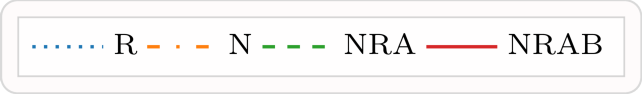}
    \end{subfigure}
    \hfill
    \vspace{-4mm}
    \caption{\textbf{Diversity measured during exploration of Gray Scott reaction-diffusion and Lenia.}
    Per each configuration, a total of $50$ run as been conducted with different random seed, the mean is depicted by curves, shaded areas represent std.
   The representation space used to measure diversity corresponds to the 13 Haralick features \cite{haralick_textural_1973} extracted from the observation, standardized and projected in a 4 dimensional space by a principal component analysis.
    The $N_{init} = 250$ first samples are chosen randomly. (a) Global diversity of Gray-Scott using $N_{bin} := 200,000$.
    (b) Constrained diversity of Gray-Scott using $N_{bin} := 100,000$.
    (c) Global diversity of Lenia using $N_{bin} := 200,000$.
    (d) Constrained diversity of Lenia using $N_{bin} := 100,000$.
    We use the following parameters. 
    For (a,b): $f \in [0.001, 0.2]$, $\sigma_m(f)=0.2$; $k \in [0.01, 0.075]$, $\sigma_m(k)=0.001$.
    For (c,d):
    $R \in [2, 40]$;
    $T = \frac{1}{dt} \in [2, 20]$;
    $\mu_k \in [0.05, 0.5]$;
    $\sigma_k \in [0.001, 0.18]$;
    $h_k \in [0.01, 1]$;
    $r_k \in [0.2, 1]$;
    $b_i \in [0.001, 1]$;
    $w_i \in [0.01, 0.5]$;
    $a_i \in [0, 1]$.
    All $\sigma_m$ are equal to $0.2$, except $\sigma_m(T)=0.5$ and $\sigma_m(\sigma_k)=0.01$.}
    \label{fig:plots}
    \vspace{-2mm}
\end{figure*}

\subsection{Evaluation}

We rely on acceptance and diversity metrics to compare our exploration approach (NRAB) to the three alternatives methods (R, N and NRA).
When randomly sampling the parameter space, 90.8\% of Gray-Scott samples lead to a homogeneous pattern (i.e., a homogeneous distribution of the same pixel value), while this only occurs for 19.0\% of samples with Lenia.
Consequently, as both systems use the same number of initial samples ($N_{init} = 250$), the history with which IMGEP methods are initialized is more diverse for Lenia than for Gray-Scott.

We next measure the proportion of samples leading to inlier classifications of patterns, which we call the acceptance rate.
The higher the rate, the better the method is able to produce samples belonging to the ROI.
Results are given in Table~\ref{tab:nbsamples} for both studied systems and all four exploration methods, without taking the initial random samples into account.
Despite the low initial diversity observed with Gray-Scott, our method (NRAB) achieves an acceptance rate $15\times$ to $20\times$ higher than that of R and approximately $2.4\times$ higher than N and NRA.
The acceptance rate of NRAB on Lenia remains the highest, even though less so ($2\times$ to $2.3\times$ higher than R, $1.6\times$ higher than N or NRA).
In both systems, given the time required to evaluate a sample, the NRAB method finds samples belonging to the ROI more quickly.

\begin{table*}[h!]
\center
    \vspace{-2mm}
    \caption {\textbf{Acceptance rates.}
    For each exploration method : the percentage of valid samples, the time required to evaluate a sample and the time required to obtain a valid sample are given (Measurements are taken with Intel Core i7-4790K CPU and Nvidia Titan V GPU).}
    \begin{tabular}{l|cccc||cccc|}
\cline{2-9}
                                       & \multicolumn{4}{c||}{\textbf{Gray-Scott}}                                                                                 & \multicolumn{4}{c|}{\textbf{Lenia}}                                                                                        \\ \cline{2-9} 
                                       & \multicolumn{1}{c|}{R}            & \multicolumn{1}{c|}{N}      & \multicolumn{1}{c|}{NRA}   & \multicolumn{1}{c||}{NRAB} & \multicolumn{1}{c|}{R}            & \multicolumn{1}{c|}{N}      & \multicolumn{1}{c|}{NRA}     & \multicolumn{1}{c|}{NRAB} \\ \hline
\multicolumn{1}{|l|}{Acceptance rate}  & \multicolumn{1}{c|}{0.68\%}       & \multicolumn{1}{c|}{5.35\%} & \multicolumn{1}{c|}{5.6\%} & \textbf{13.66\%}          & \multicolumn{1}{c|}{8.33\%}       & \multicolumn{1}{c|}{11.7\%} & \multicolumn{1}{c|}{10.99\%} & \textbf{19.46\%}          \\ \hline
\multicolumn{1}{|l|}{Time/sample (ms)} & \multicolumn{1}{c|}{\textbf{155}} & \multicolumn{1}{c|}{353}    & \multicolumn{1}{c|}{388}   & 412                       & \multicolumn{1}{c|}{\textbf{767}} & \multicolumn{1}{c|}{796}    & \multicolumn{1}{c|}{823}     & 845                       \\ \hline
\multicolumn{1}{|l|}{Time/inlier (s)}  & \multicolumn{1}{c|}{22.784}       & \multicolumn{1}{c|}{6.598}  & \multicolumn{1}{c|}{6.928} & \textbf{3.016}            & \multicolumn{1}{c|}{9.207}        & \multicolumn{1}{c|}{6.803}  & \multicolumn{1}{c|}{7.488}   & \textbf{4.342}            \\ \hline
\end{tabular}
    
    \label{tab:nbsamples}
    \vspace{-3mm}
\end{table*}

To measure global diversity, we choose a metric already used in quality-diversity search \cite{pugh_confronting_2015}.
It works by partitioning the representation space into $N$ bins.
Each observation is embedded in the associated bin of the behavior space. 
Diversity then corresponds to the number of bins in which at least one observation is embedded.
The constrained diversity metrics is computed the same way, except that we \emph{only consider inliers}.
Results are shown in Figure~\ref{fig:plots} for both types of diversity and the two studied systems.
A qualitative assessment of these results is shown in Figure~\ref{fig:patterns_preview}.

As expected, IMGEP methods perform better than random parameter sampling.
NRA leads to greater global diversity than N, which is expected given the variance in the distribution of behaviors along the 10 axes.
The global diversity of NRAB is very close to that of NRA, thus balanced sampling allows for global diversity to be maintained.
IMGEP methods appear to be less effective on Lenia. As mentioned above, the rate of homogeneous patterns produced by uniform sampling of Lenia's parameter space is low.
Furthermore, compared with Gray-Scott, Lenia offers greater variation in the patterns it produces.
Thus, for a few samples, random sampling already produces a significant diversity.
The effectiveness of the IMGEP method for Lenia is more marked when a larger number of samples is employed: as a comparison, Reinke and al. and Etcheverry and al. \cite{reinke_intrinsically_2020,etcheverry_hierarchically_2020} used 1,000 initialization samples for a total of 5,000 samples.

\begin{figure}[h!]
    \centering
    \vspace{-2mm}
    \begin{subfigure}{0.47\textwidth}
        \centering
        \subcaption{Gray-Scott}
        \includegraphics[width=\linewidth]{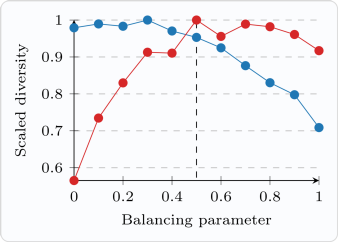}
    \end{subfigure}
    \begin{subfigure}{0.47\textwidth}
        \centering
        \subcaption{Lenia}
        \includegraphics[width=\linewidth]{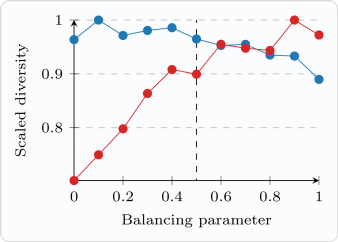}
    \end{subfigure}
    \begin{subfigure}{\textwidth}
        \centering
        \includegraphics[width=0.5\linewidth]{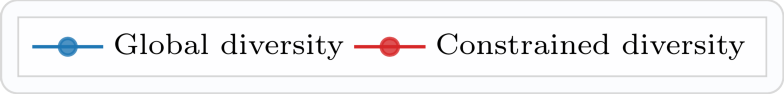}
    \end{subfigure}
    \vspace{-2mm}
    \caption{\textbf{Study of the balancing meta-parameter.} The conditions are similar to those described in the Section~\ref{sec:results}. Each point represents the diversity after $1000$ samples using NRAB method, divided by the maximum diversity encountered, for (a) Gray-Scott and (b) Lenia.}
    \label{fig:meta_param}
    \vspace{-6mm}
\end{figure}

Most importantly, our method achieves greater constrained diversity than the others.
The effectiveness of balanced sampling is demonstrated by the increase in constrained diversity of NRAB compared to NRA, for both Gray-Scott and Lenia.
We further show in Figure~\ref{fig:meta_param} the impact of the balancing meta-parameter on constrained diversity.
A value of $1$ (only sampling the ROI) yields high constrained diversity but low global diversity; whereas a value of $0$ (sampling the whole space) yields the opposite situation, which is prone to local minima issues.
Our choice of a rate of $0.5$ thus appears as a safe compromise.

\section{Discussion}
\label{sec:discuss}
In this paper, we have considered ways to incorporate interactive human guidance in complex system exploration.
In practice, we employ a modified IMGEP, augmenting the history to accommodate for variable user ROI.
We use a balanced policy to sample the parameter space, which maintains global diversity while achieving better constrained diversity and acceptance rates.
As shown in our experiments on Gray-Scott and Lenia reaction-diffusion, our method achieves better results while remaining agnostic to the mapping between system parameters and behaviors.
We now discuss each of our design choices in turn.

The use of explicit constraints (\ref{choice:constraints}) allows for the ROI to be changed without losing information from past samples, as all samples lead to a greater global diversity.
However, not all regions of interest can be formulated through intuitive explicit constraints.
To get around this problem, we might consider using implicit constraints (i.e., labelling example inliers/outliers from exploration samples).

As shown in Figure~\ref{fig:meta_param}, maintaining global diversity (\ref{choice:diversity}) is beneficial.
However, the choice of balance meta-parameter depends on the user ROI and considered system.
A dynamic balancing based on statistical measurements or user feedback could be explored in future work.
It is also possible to use a phase space approach to adapt the parameter sampling policy. 
As shown by Papadopoulos \cite{papadopoulos_looking_2024}, phase transition zones tend to bring greater diversity, which is beneficial for global diversity but needs to be reconsidered in the case of constrained diversity.

The definition of a non-specific behavior space (\ref{choice:behavior-space}) eliminates the need for prior knowledge of the system outputs.
However, the exploration process could be augmented with priors without altering the overall behavior.
When available, apriori knowledge of the system behavior could lead to the construction of a more effective behavior space that embeds a greater diversity (e.g., more initialization samples to get around Gray-Scott's low acceptance rate).

In future work, we would like to consider two extensions of our approach.
We chose not to incorporate the initial state in the search space, in order to only focus on the dynamic behavior of the system as a function of the update rule.
However, the methods employed could be used to explore all the basins of attraction in behavior space of a complex system for a fixed update rule.
Furthermore, this work is focused on systems producing image/texture output, but it is possible to extend the scope of the considered systems to those producing any other type of output. In particular, those whose observation corresponds to system dynamics (i.e., $\obsmapping(R_{0\rightarrow T}(\mathcal{S}, \param_i)) := R_{0\rightarrow T}(\mathcal{S}, \param_i)$ ), such as swarm/flock/crowd models \cite{reynolds_flocks_1987}.

\footnotesize
\bibliographystyle{splncs04}
\bibliography{main}

\end{document}